\pdfoutput=1
\documentclass{article}
\usepackage{ml4cps}
\usepackage{graphicx} 
\usepackage[utf8]{inputenc} 
\usepackage[T1]{fontenc}    
\usepackage{hyperref}       
\usepackage{url}            
\usepackage{booktabs}       
\usepackage{amsfonts}       
\usepackage{nicefrac}       
\usepackage{microtype}      
\usepackage{cleveref}       
\usepackage{indentfirst}
\usepackage{parskip}
\usepackage{xcolor}
\UseRawInputEncoding
\title{Combining UPerNet and ConvNeXt for Contrails Identification to reduce Global Warming}

\date{}

\newif\ifuniqueAffiliation
\uniqueAffiliationtrue

\ifuniqueAffiliation 
\author{ {Zhenkuan Wang}\\
    \texttt{xiaowang0662@gmail.com} \\
}

\begin{document}
    \maketitle
    \begin{abstract}
Semantic segmentation is a critical tool in computer vision, applied in various domains like autonomous driving and medical imaging. This study focuses on aircraft contrail detection in global satellite images to improve contrail models and mitigate their impact on climate change.An innovative data preprocessing technique for NOAA GOES-16 satellite images is developed, using brightness temperature data from the infrared channel to create false-color images, enhancing model perception. To tackle class imbalance, the training dataset exclusively includes images with positive contrail labels.The model selection is based on the UPerNet architecture, implemented using the MMsegmentation library, with the integration of two ConvNeXt configurations for improved performance. Cross-entropy loss with positive class weights enhances contrail recognition. Fine-tuning employs the AdamW optimizer with a learning rate of $2.5 \times 10^{-4}$.During inference, a multi-model prediction fusion strategy and a contrail determination threshold of 0.75 yield a binary prediction mask. RLE encoding is used for efficient prediction result organization.The approach achieves exceptional results, boasting a high Dice coefficient score, placing it in the top 5\% of participating teams. This underscores the innovative nature of the segmentation model and its potential for enhanced contrail recognition in satellite imagery.For further exploration, the code and models are available on GitHub:  \url{https://github.com/biluko/2023GRIC.git}.

    \end{abstract}


    \section{Introduction}
    \subsection{Background and Motivation}
\begin{figure}[ht]
        \centering
        \includegraphics[width=1\textwidth]{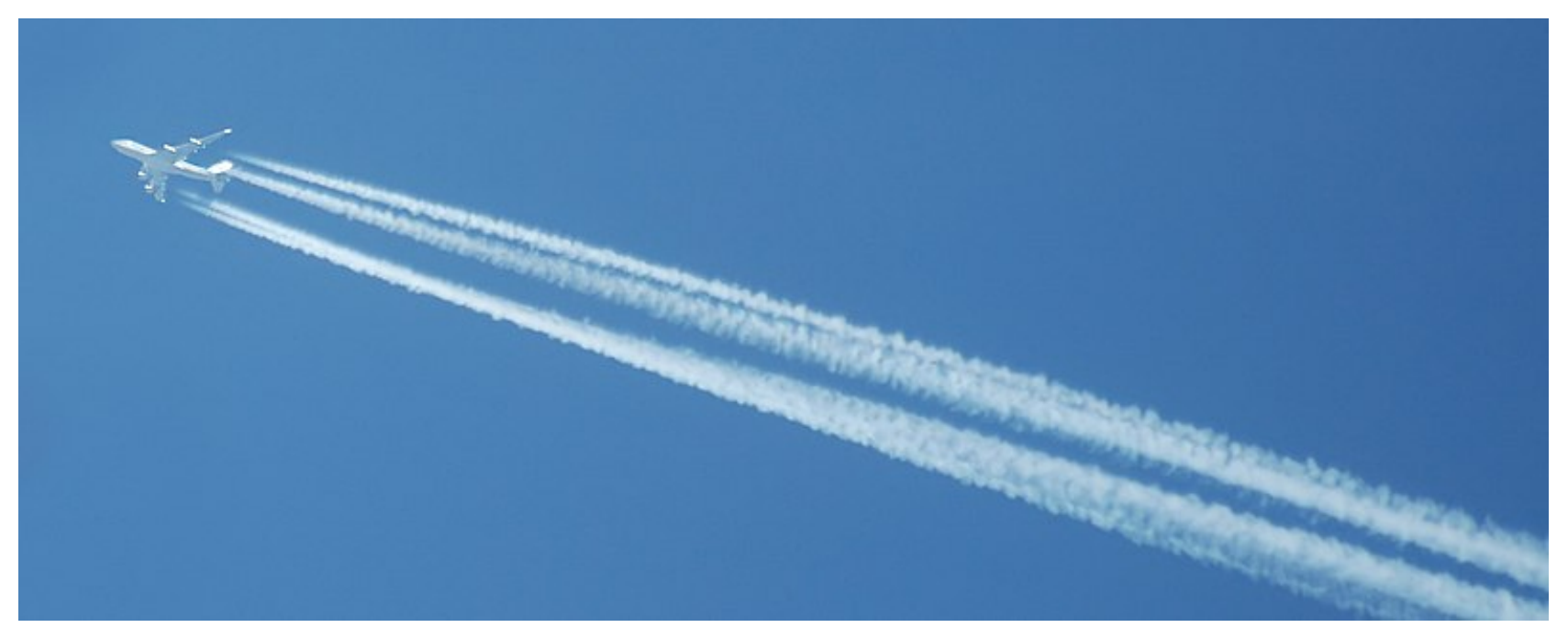}
        \caption{Contrail with jet.Image from:\url{https://commons.m.wikimedia.org/wiki/File:Contrail_with_jet_(aka).jpg}}
        \label{fig:1}
    \end{figure}

    Contrail avoidance stands out as one of the most scalable, cost-effective, and environmentally sustainable solutions available to the aviation industry today.Contrails, which are linear clouds of ice crystals formed at high altitudes due to aircraft engine emissions in ultra-humid atmospheres, have gained increasing attention due to their contribution to global warming by trapping heat in the atmosphere.From the Figure \ref{fig:1}, it can be observed\cite{jensen1998environmental,mannstein2005note}. In fact, research has revealed that the warming effect caused by contrails in global aviation is comparable to that of carbon dioxide emissions\cite{lee2021contribution,bickel2020estimating,avila2019reducing,changnon1980effect}.

    The phenomenon of contrails was initially observed by military aircraft approximately 75 years ago, with the initial objective of minimizing their visible presence in the sky. About three decades ago, climate scientists in Europe began recognizing that contrails play a role in blocking the Earth's normal nighttime heat release. They developed sophisticated models based on weather data to predict the timing and extent of contrail formation. These models' predictions have since been validated by various research laboratories, establishing that contrails contribute to approximately 1\% of human-induced global warming\cite{sassen1997contrail,lee2000identifying}.

    The advancement of this research has increasingly relied on satellite imagery to validate the predictions made by these models. The use of such imagery provides robust validation, instilling greater confidence in these models among pilots. Furthermore, the aviation industry gains a reliable means to assess the effectiveness of contrail avoidance strategies.

\begin{figure}[ht]
        \centering
        \includegraphics[width=0.7\textwidth]{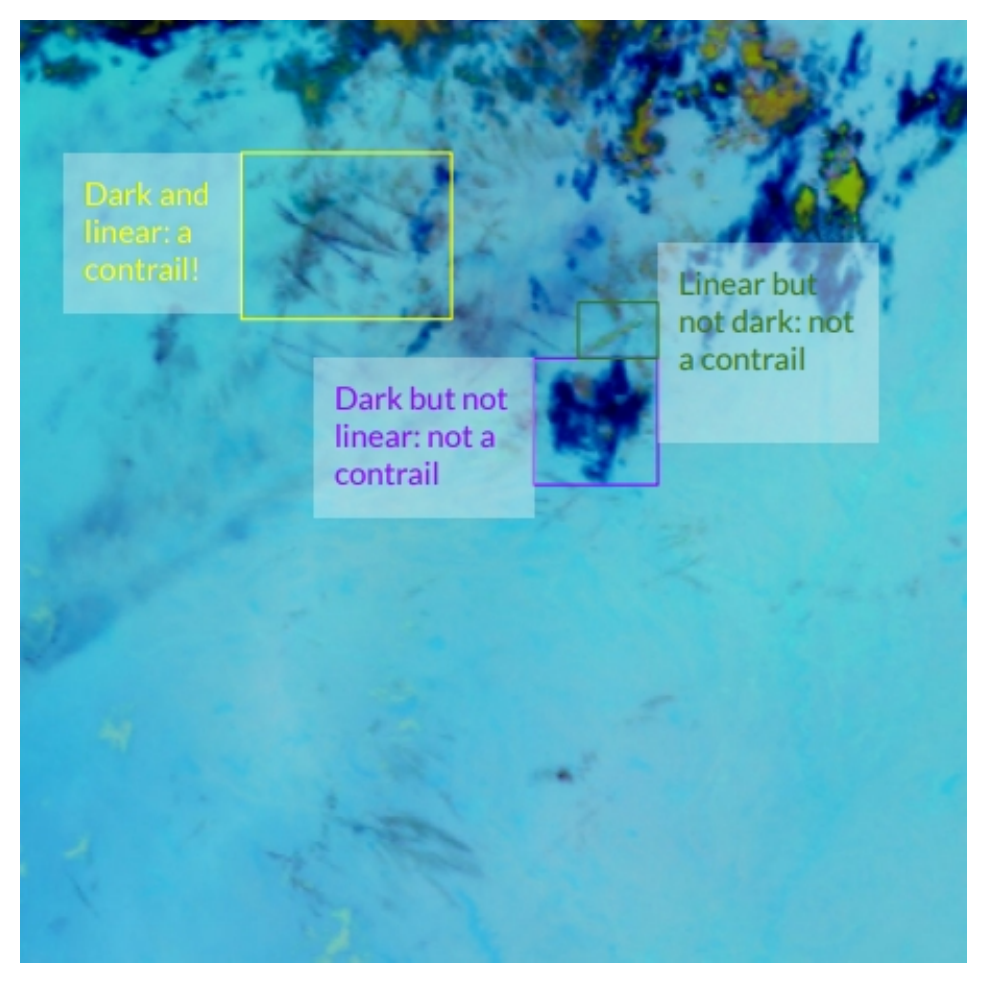}
        \caption{The image is calibrated so that contrails will appear dark in color.An object is a contrail if it is both line-shaped and darker than its surroundings.}
        \label{fig:2}
    \end{figure}

    Detecting contrails in satellite images is a challenging task because they visually resemble natural cirrus clouds\cite{dekoutsidis2023contrail,mannstein2012contrail,mannstein2010ground,minnis2011estimating,minnis2013linear}. Contrail formations manifest as linear ice clouds that gradually change shape over time, making them difficult to differentiate from their natural cirrus counterparts. While computer vision has made significant progress with the advent of large-scale datasets like ImageNet and neural networks\cite{krizhevsky2017imagenet,smirnov2014comparison}, recognizing domain-specific issues such as those posed by infrared satellite images remains a challenge, primarily due to the scarcity of relevant training data. In contrast to traditional object recognition tasks, where texture and color cues are abundant, contrails often exhibit visual similarities, or even indistinguishability, from natural cirrus clouds. Furthermore, human experts must consider not only color, texture, and linear shape but also the temporal evolution of contrails, as they form rapidly and expand over time, unlike natural cirrus clouds.The Figure \ref{fig:2} is a false-color image of some infrared satellite bands\cite{attachoo2009new,yin2012false}.

    In this study, we leverage a dataset meticulously labeled by human experts, offering comprehensive spatiotemporal coverage. Each image in this dataset\cite{ng2023opencontrails} includes a high-quality per-pixel human-labeled trajectory mask for contrail detection. Initially, we train a Unet model\cite{ronneberger2015u} for the purpose of contrail detection and demonstrate that the resulting model produces highly accurate contrail detection outputs. We envision this as a benchmark for future track detection tasks and aspire to expedite progress in this field by promoting the widespread adoption of large-scale contrail detection results.
    \subsection{Segmentation}
    Image segmentation\cite{minaee2021image,haralick1985image,zhang1996survey,pal1993review} is a paramount subject in the realm of image processing and computer vision, finding extensive applications in various domains like scene comprehension, medical image analysis, robot perception, video surveillance, augmented reality, and image compression. In recent years, thanks to the remarkable performance of deep learning models\cite{campos2016evaluation,zhou2017application,he2016deep,lecun2015deep,tan2019efficientnet,chen2014semantic,zhang2022resnest} in visual tasks, there has been a surge in research endeavors aimed at innovating image segmentation methods through the integration of deep learning techniques.

    Image segmentation can be broadly categorized into two primary types: semantic segmentation and instance segmentation. Semantic segmentation involves the classification of pixels into semantically meaningful labels, such as identifying objects like people, cars, and chairs within an image. This task is notably more complex than standard image classification, which assigns a single label to the entire image. On the other hand, instance segmentation takes this a step further by not only detecting objects of interest but also precisely delineating each individual object, thereby enabling them to be distinctly outlined within the image.

    In the realm of computer vision, these advancements in image segmentation are driving progress and innovation, empowering applications across a wide spectrum of fields.

   Since its ascent around 2015, deep learning has found applications across a myriad of domains, with image segmentation being no exception. Numerous architectural innovations have emerged for tackling segmentation tasks, including well-known approaches such as FCN, UNet, DeepLab, and Transformer-based models\cite{simonyan2014very,long2015fully,chen2018encoder,chen2017deeplab}.

   Fully Convolutional Networks (FCN) were among the pioneering architectures, introducing a single upsampling convolutional layer towards the end to generate pixel-wise segmentation masks. U-Net, on the other hand, elegantly addressed the vanishing gradient problem by progressively upsampling the feature map of the residual connection during backpropagation. This development led to a proliferation of encoder-decoder variants. However, the skip-connections in U-Net often exhibited rough transitions, resulting in substantial semantic differences between the inputs of connected convolutional layers, posing challenges for the network's learning process. To mitigate this semantic gap, U-Net++\cite{zhou2018unet++} was introduced, enhancing the U-Net's direct connections with convolutional layers reminiscent of Dense structures, effectively integrating features from subsequent convolutional stages. Additionally, earlier semantic segmentation networks relied on pooling, causing information loss and neglecting label probabilistic relationships. The DeepLab series addressed these issues by employing atrous (dilated) convolutions to circumvent information loss arising from pooling.

   The Transformer model, initially devised as an attention mechanism, initially found wide application in sequential data processing realms like speech recognition and natural language processing, enabling the learning of global contextual information from input sequences. In recent years, this concept has transcended into visual signal processing, yielding remarkable results. The fundamental idea involves dividing the image into discrete blocks, transforming them into a sequence, and subsequently applying an attention mechanism akin to that used in sequential data.

   Within this framework, several impressive architectures have emerged. Vision Transformer (ViT) \cite{dosovitskiy2020image}notably introduced a pure Transformer architecture, eliminating convolutional layers entirely, yet outperforming ResNet152 on various prominent datasets. Swin introduced the concept of shifting windows to generate patches, facilitating information exchange between patches. The Unet and Swin concepts were amalgamated to form a U-shaped encoder-decoder structure, echoed in UNetR and TransUnet. On a different note, BEiT borrowed the notion of mask tagging from the BERT language model, adapting it to the image Transformer, and employed automatic encoding to tag individual image patches for masking.

   These advancements have profoundly enriched the realm of image processing, providing more efficient and accurate tools, propelling visual tasks into a significant frontier driven by the advances in deep learning.
    \subsection{Dataset}
    I will utilize geostationary satellite imagery to detect aviation contrails. The raw satellite imagery has been sourced from the GOES-16 Advanced Baseline Imager (ABI), which is publicly accessible on Google Cloud Storage. The ABI, a remote sensing instrument developed by NOAA in the United States, is deployed on a geostationary satellite to capture high-resolution data of the Earth's atmosphere and weather conditions. It continuously monitors weather patterns in North America and the Atlantic region.

    The GOES-16 ABI achieves global and local weather imaging through a range of multispectral channels, including visible light, infrared, and ultraviolet, enabling the monitoring of clouds, temperature, ocean conditions, and more. Its high resolution and real-time capabilities provide invaluable data for weather forecasting, meteorological monitoring, and early warnings of natural disasters.

    While the nominal pixel size of the GOES-16 ABI is as small as $2 \times 2$ kilometers, rendering the initial stages of contrail formation nearly invisible, it's important to note that the warming effect of contrails predominates during the extended hours they persist. Therefore, the absence of short-lived contrails is unlikely to significantly impact our assessment or our ability to mitigate the warming caused by most persistent contrails.

\begin{figure}[ht]
        \centering
        \includegraphics[width=0.7\textwidth]{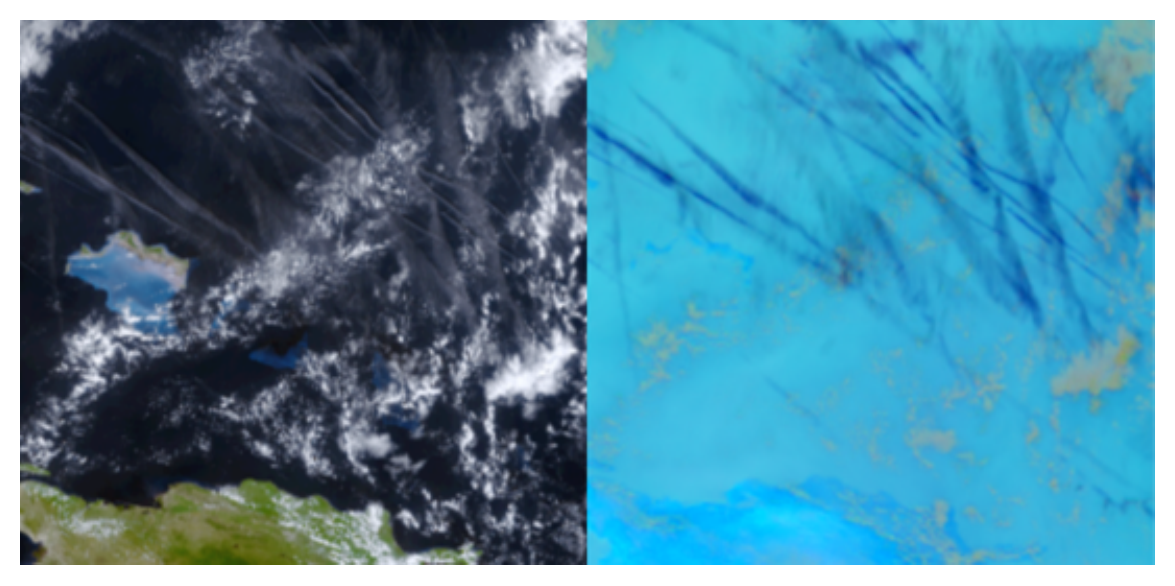}
        \caption{Real satellite images and False color images.}
        \label{fig:3}
    \end{figure}

    To enable continuous contrails detection around the clock, Joe et al\cite{ng2023opencontrails}. employed a "grey" false color scheme for image representation. This scheme amalgamates data from three long-wavelength GOES-16 brightness and temperature channels: the red channel corresponds to 12$\mu m$, the blue channel depicts the difference between 12$\mu m$ and 11$\mu m$, and the green channel represents 11$\mu m$ minus the difference from 8$\mu m$. This color scheme has been specifically designed to enhance the visibility of ice clouds and facilitate the identification of contrails.As shown in Figure \ref{fig:3}, the false color scheme is used for image representation.

\begin{figure}[ht]
    \centering
    \begin{minipage}{0.49\textwidth}
        \centering
        \includegraphics[width=\textwidth]{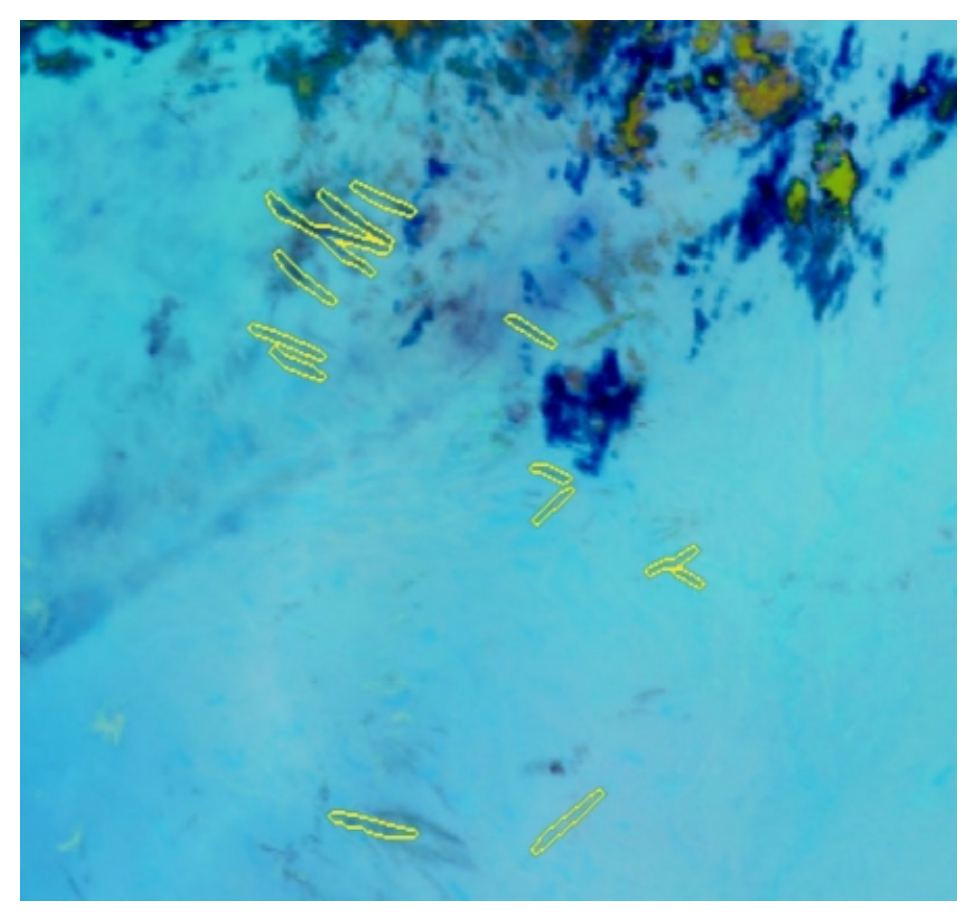}
    \end{minipage}
    \hfill
    \begin{minipage}{0.49\textwidth}
        \centering
        \includegraphics[width=\textwidth]{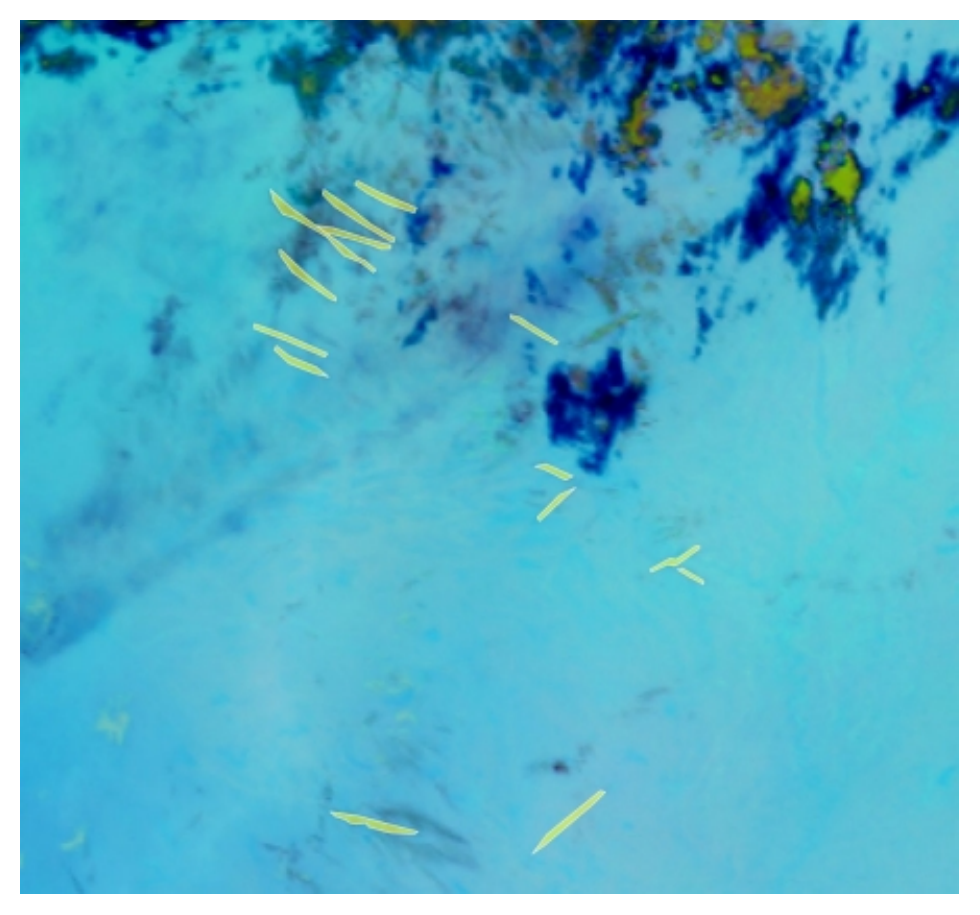}
    \end{minipage}
    \caption{Contrails Labeling. Left: here the objects which are contrails have been highlighted. Right: this is what the image should look like after you are done labeling it.}
    \label{fig:6}
\end{figure}

    Each image patch within the dataset corresponds to an approximate geographical area of $500 \times 500$ km. The original full-disk images are subjected to bilinear resampling to create localized scene images. Annotators are presented with image patches sized at $281\times 281$ pixels. It's worth noting that labels near boundaries might exhibit noise due to the absence of contextual information.The polygon should include all of
    the dark pixels that make up the contrail,as shown in Figure \ref{fig:6}.

    For both training and evaluation purposes, the decision was made to crop images and contrails masks to dimensions of $256 \times 256$ pixels. Recognizing that contrails identification benefits from temporal context, a sequence of images, captured at 10-minute intervals, is made available. Multiple GOES-16 ABI brightness temperature values and brightness temperature differences are provided for each example.

    This revised description offers a more professionally articulated explanation of the methods and data used in the contrails detection process.

    While prior research has made attempts to employ deep convolutional neural networks for training contrails detection models, the performance of these models still lags significantly behind human experts\cite{zhang2018contrail,meyer2002regional}. Consequently, there remains ample room for enhancing model performance. One plausible explanation for this gap is that human experts can comprehensively consider the temporal evolution of contrails during the labeling process, whereas existing computer models typically operate on single frames. To address this challenge, a novel model has been developed to incorporate temporal context from multiple input frames. Experimental results have demonstrated that methods considering multiple frames have achieved notable improvements in contrails detection performance compared to models processing single frames alone.

    Within the dataset folder, each subdirectory is identified by a unique {record\_id} and contains a binary file in the numpy.npy format, corresponding to a single example.

\begin{figure}[ht]
        \centering
        \includegraphics[width=1\textwidth]{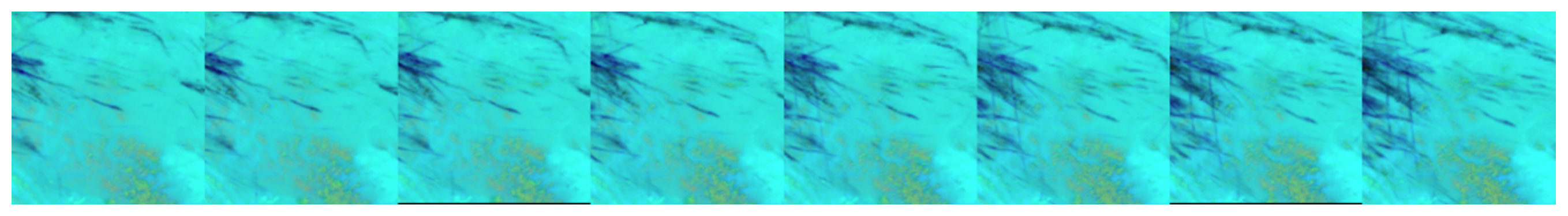}
        \caption{contrails initially appear as sharp lines that gradually disperse over time.}
        \label{fig:7}
    \end{figure}

    To facilitate accurate identification and labeling of contrails, specific guidelines have been provided. Contrails often manifest as darker linear features in satellite images when utilizing a grayscale scheme. Aircraft contrails initially appear as sharp lines that gradually disperse over time and may span multiple consecutive frames. Distinguishing contrails from static ground-based structures such as roads, rivers, or coastlines can be aided by these characteristics. Additionally, we offer comprehensive annotation guidelines to ensure accurate and consistent labeling practices.As shown in Figure \ref{fig:7}.

    This revised text conveys a more professional tone while retaining clarity in describing the research and dataset.

    \begin{enumerate}
        \item Contrails must contain at least 10 pixels.
        \item At some point during their existence, contrails must be at least 3 times longer than they are wide.
        \item Contrails must either appear suddenly or enter from the sides of the image.
        \item Contrails should be visible in at least two images.
    \end{enumerate}

    The complete dataset comprises 20,529 examples in the training set, 1,856 examples in the validation set, and 2 examples in the test set. The examples were randomly selected, with the exception of satellite scenes flagged by Google Street View as potential contrails, which were exclusively included in the training set. Notably, among the training examples, 9,283 contain at least one annotated contrail.

    When it comes to evaluating image segmentation, two widely used metrics are Intersection over Union (IoU) and the Dice score\cite{eelbode2020optimization,zhou2019iou}. IoU primarily quantifies the degree of overlap between predicted and actual segmentation results. However, it's important to note that IoU doesn't account for subtle pixel-level variations, potentially limiting its applicability in certain scenarios. In contrast, the Dice coefficient assesses pixel-level agreement between predicted segmentation and corresponding ground truth. For the foreground positive class in binary classification, the Dice score is equivalent to the F1 score, encompassing precision and recall.

    It is worth mentioning that Dice loss is often used as a loss function, namely 1-Dice Score. This form of loss function helps model optimization, focusing on maximizing the Dice score, thereby improving segmentation performance. In image segmentation tasks, IoU and Dice scores are key evaluation indicators that can help judge the performance of the model and take into account the quality and accuracy of the segmentation results. However, be aware of the limitations of these metrics, and sometimes it may be necessary to combine them with other metrics to more fully evaluate the performance of the segmentation model.
    $$
    Dice(A, B) = \frac{2|A \cap B|}{A + B} = \frac{2TP}{2TP + FP + FN} = F1
    $$

    \section{Contrails Iamge Segmentation}
    In prior research, the prevalent approach for wake detection involved the use of convolutional neural network models. These models would produce a confidence score, ranging from 0 to 1, for each pixel, signifying the likelihood that the pixel belongs to a wake trajectory. Subsequently, post-processing of the wake detection output at each pixel would be performed to generate line segments representing instances of wake trajectories. To ensure the consistency of wake detection across diverse scenarios, including nighttime conditions, the choice was made to train the model using the infrared channel with brightness temperature as the input feature.

    The established wake detection model takes eight channels as input, specifically 8$\mu m$, 10$\mu m$, 11$\mu m$, 12$\mu m$, as well as the differences between 12$\mu m$ and 11$\mu m$, and between 11$\mu m$ and 8$\mu m$. Prior to inputting this data into the network, a normalization step is applied. This normalization involves subtracting the global mean and dividing each channel by the channel's global variance. This preprocessing method is employed to ensure that the input data maintains a certain level of stability and comparability throughout the training process.
    \subsection{Established Image Segmentation Model}
    In the field of aerial imaging, UNet and DeepLab are among the most widely used models because they perform well in image segmentation tasks and can efficiently consider global and local features. These models are able to meticulously handle minute details in imaging while also capturing global background information.

    Although both U-Net and DeepLab are used for image segmentation, their architectures are very different. UNet adopts an encoder-decoder structure to pass feature maps between different levels through skip connections to achieve segmentation. Different from this, DeepLab uses attribute convolution, which allows to build contextual information at multiple feature levels. Although attribute convolution is very useful for capturing multi-level features, it also causes DeepLab to require more computing resources during the training process, making it a more computationally expensive model. Therefore, in our project, we chose to use UNet because it does not involve additional calculations and can effectively solve the vanishing gradient problem.
    \subsection{Recent Advances in Segmentation Techniques}
    Presently, fully convolutional networks (FCN)\cite{long2015fully} based segmentation networks are widely regarded as the preferred choice. Due to the typically limited availability of ample training data in segmentation tasks, it is common practice to initialize networks with models pretrained for image classification tasks. To attain high-resolution predictions in semantic segmentation, dilated convolution technology has been introduced. This technology effectively enlarges the receptive field range while alleviating the adverse consequences of downsampling. Networks employing dilated convolution technology have evolved into a standard framework for semantic segmentation tasks. However, it's important to acknowledge that this paradigm is not without its limitations:

    \begin{enumerate}

   \item While deep convolutional networks, as recently introduced, have enjoyed remarkable success in image classification and semantic segmentation tasks, they frequently comprise tens or even hundreds of layers.

    \item The architectural complexity of these networks has led to a rapid upsurge in downsampling during the initial stages of the network, owing to the inclusion of larger receptive fields and reduced computational complexity.

   \end{enumerate}

    Typically, such networks primarily exploit the deepest feature maps. While in segmentation tasks, it's logical to employ high-level semantic features for segmenting higher-level concepts (e.g., objects), this approach may not be well-suited for segmenting multi-level perceptual attributes, particularly lower-level concepts (e.g., texture, material).

\begin{figure}[ht]
        \centering
        \includegraphics[width=1\textwidth]{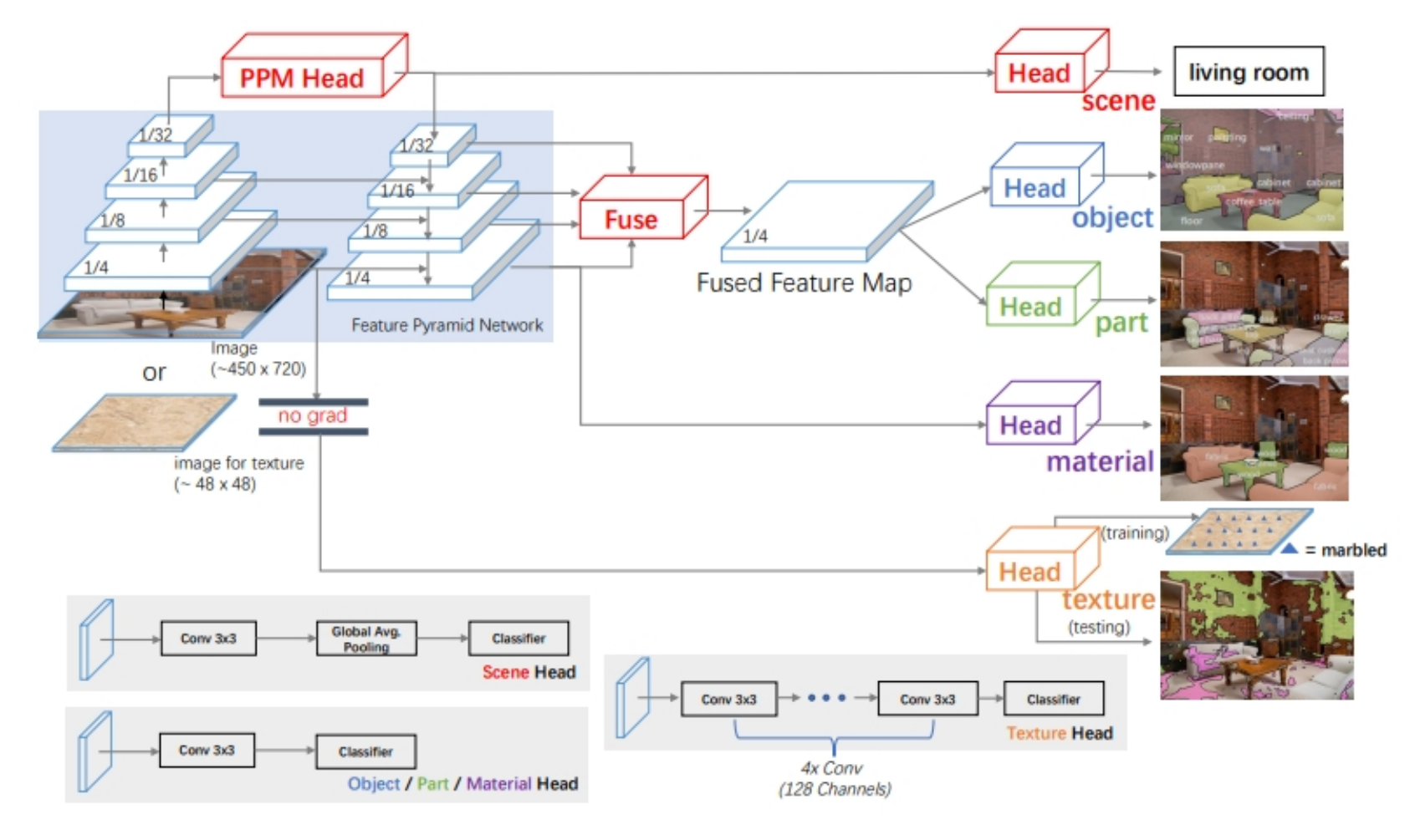}
        \caption{UPerNet framework for Unified Perceptual Parsing.The architecture is from \cite{xiao2018unified}.}
        \label{fig:8}
    \end{figure}

    The UPerNet (Unified Perceptual Parsing Network)\cite{xiao2018unified} is constructed upon the foundation of the Feature Pyramid Network (FPN)\cite{lin2017feature}. Despite the theoretically wide receptive fields of deep convolutional networks, their practical usable receptive fields tend to be much smaller. To address this challenge, the Pyramid Pooling Module (PPM), as seen in PSPNet\cite{zhu2021coronary}, is employed at the final layer of the backbone network. Subsequently, the output from PPM is passed on to the top-down FPN branch.As shown in Figure \ref{fig:8}.

    This dataset incorporates features at multiple semantic levels. Given that image-level information is more suited for scene classification, the Scene head is directly connected to the feature map following the PPM module. The Object head and Part head are attached to the feature map that integrates all layers from FPN. The Material head is linked to the FPN feature map with the highest resolution. Texture is associated with the Res-2 module in ResNet and is fine-tuned after the entire network completes training on other tasks.

    This design is driven by three key considerations:

    \begin{enumerate}
        \item Texture represents the most fundamental perceptual attribute, allowing predictions to be made based solely on salient features without relying on high-level information.
        \item While training on other tasks, the network implicitly learns the critical features necessary for accurate texture predictions.
        \item The receptive field of this branch must be sufficiently small to enable distinct labels for different regions when processing a standard-sized image.
    \end{enumerate}

    \section{Methodology and Experiments}
    In my approach, a series of five experiments were conducted, each tailored to uncover a distinct facet of the deep learning model. Building upon the foundation of UNet, various encoders including ResNet and others were incorporated to assess the Dice coefficient performance on an imbalanced dataset. Subsequently, different network architectures were evaluated, culminating in the introduction of UpperNet in conjunction with the ConvNeXt Network\cite{liu2022convnet} to enhance overall performance.

    \subsection{Data Preprocessing}
    The core idea of the approach is to enable contrails detection in both daytime and nighttime conditions by presenting images to human annotators in a "gray" false color scheme. This scheme combines three GOES-16 brightness temperature channels: 12$\mu m$, the difference between 12$\mu m$ and 11$\mu m$, and the difference between 11$\mu m$ and 8$\mu m$, with the aim of highlighting ice clouds, including contrails, as dark features. Each image patch in the dataset corresponds to an area of approximately $500 \times 500$ km and is reprojected to the Universal Transverse Mercator (UTM) coordinate system using bilinear resampling. An image patch of size $281\times 281$ is presented to taggers, and labels near boundaries, often noisy due to the absence of spatial context, are subsequently cropped to $256 \times 256$ pixels for both the image and contrails masks during training and evaluation.

\begin{table}[]
    \centering
    \caption{Ash RGB}
    \label{tab:ash_rgb}
    \begin{tabular}{|c|c|ccc|c|}
        \hline
        Color Beam & Channel       & \multicolumn{3}{c|}{Range}                              & Gamma \\ \hline
        Red        & IR12.0-IR10.8 & \multicolumn{1}{c|}{-4}  & \multicolumn{1}{c|}{+2}  & K & 1     \\ \hline
        Green      & IR10.8-IR8.7  & \multicolumn{1}{c|}{-4}  & \multicolumn{1}{c|}{+5}  & K & 1     \\ \hline
        Blue       & IR10.8        & \multicolumn{1}{c|}{243} & \multicolumn{1}{c|}{303} & K & 1     \\ \hline
    \end{tabular}
\end{table}

   The Table \ref{tab:ash_rgb} presents the visible channels or channel differences within the red, green, and blue beams. These measurements should initially undergo calibration to convert them into brightness temperatures. Subsequently, the channel or channel differences need to be enhanced, undergoing linear stretching within the brightness temperature range as detailed in the table.

   To achieve this, the data was initially normalized, mapping its range to the [0, 1] interval. This normalization step serves the purpose of ensuring that data from different channels shares the same scale, facilitating subsequent visualization and analysis.

   Following the initial steps, numerical values for the three channels were computed using data from various spectral bands:

    \textbf{Red Channel (r-channel):} Temperature differences were captured by calculating "band\_15 - band\_14" and then normalizing it to the (-4, 2) range.

    \textbf{Green Channel (g-channel):} Cloud-top temperature differences were captured by calculating "band\_14 - band\_11" and then normalizing it to the (-4, 5) range.

    \textbf{Blue Channel (b-channel):} Temperature information was captured by normalizing "band\_14" data to the (243, 303) range.

    Subsequently, the data from these three channels were merged to generate a false-color image, which effectively visually conveys the information captured by the different channels. In this image, distinct colors in the channels symbolize various types of information. To ensure the integrity of the image data, the pixel values were ultimately constrained within the [0, 1] range.

    This processing workflow serves as a valuable tool for visualizing remote sensing data, enabling researchers to gain clearer insights into surface temperature variations, cloud characteristics, and other relevant information. It offers robust support for subsequent analysis and research endeavors.

    Our partial training images is shown as Figure \ref{fig:9}:

        \begin{figure}[ht]
        \centering
        \includegraphics[width=1\textwidth]{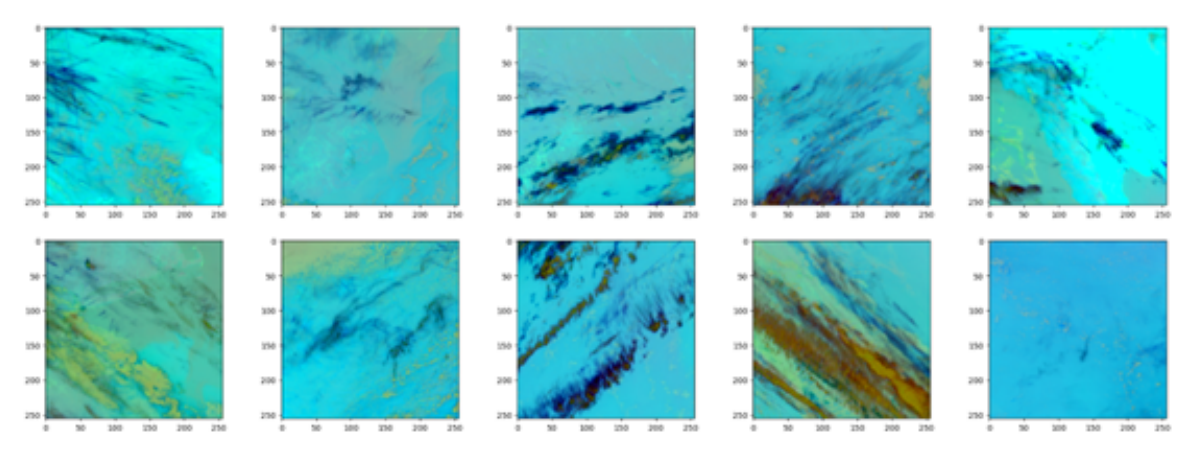}
        \caption{Example images.}
        \label{fig:9}
    \end{figure}

    \subsection{Experiment 1: Baseline UNet}
    In the initial experiments, I established a baseline model to serve as a reference point for subsequent comparisons and evaluations. To create masks for each image, annotations were extracted, particularly for those classified as "contrails." Initially, these masks were initialized as zero-padded arrays and were subsequently filled in based on the relevant annotations after validation. These masks were then transformed to match the required input dimensions of the model architecture. This meticulous process ensured that the dataset only contained images with corresponding labels, effectively incorporating filtering mechanisms to ensure that my model learned solely from images with proper mask annotations while avoiding the acquisition of irrelevant annotation patterns.

    For the baseline model, I made modifications to the UNet architecture using the SMP library\url{https://github.com/qubvel/segmentation_models.pytorch}, an enhanced version of the original UNet design. One notable enhancement involved utilizing ResNet as the encoder, endowing the model with potent feature extraction capabilities. The incorporation of ResNet's residual connections helped alleviate the challenge of vanishing gradients, which is pivotal during the training phase. Specifically, the baseline model employed ResNet50 as the encoder for the UNet.

    Moreover, the SMP UNet model offers significant flexibility, permitting straightforward exploration of specific model optimizations. The model was initialized with the Dice coefficient, and prior to the training phase, an exploration stage was conducted to determine the optimal learning rate and image dimensions for model training. Subsequently, the model underwent training for 25 epochs, with a batch size of 16, a learning rate of 0.001, and image dimensions of $384 \times 384$ being selected as the optimal configurations.

    Following the training phase, the model's performance was further assessed through visual analysis. Loss value curves were plotted to monitor the model's learning progress, detect potential overfitting or underfitting issues, and gain insights into areas that might require improvement. The Dice coefficient, specifically for the baseline model, was calculated to be 0.60034.

    \subsection{Experiment 2: Different encoder to U-Net}
    In the second experiment of the study, the focus was on exploring the potential benefits of pretraining in deep learning models. In this approach, different ResNet models were utilized, and their weights, pretrained through previous training, were employed for initialization. The objective was to determine whether this strategy could expedite the training process and enhance the overall model performance. The evaluation metric employed was the Dice coefficient, and the application of pretraining techniques yielded a notable improvement compared to the baseline model. This compelling evidence suggests that pretraining has a positive impact on enhancing the model's performance in satellite image segmentation tasks.

    One of the primary advantages of pretraining, particularly beneficial for relatively small datasets like those in medical imaging, is its capability to mitigate the risk of overfitting. Overfitting tends to occur when a model becomes overly complex and starts to overadapt to the features of the training data, rather than extracting generalizable features. By utilizing diverse pretrained models such as ResNet, ResNext, and ResNest, trained on large-scale datasets, a broad spectrum of prelearned features can be leveraged. These features span from basic structures like edge detection at the outset to more intricate shapes at deeper layers. This inherent advantage of pretraining contributes to a more effective and efficient learning process, enabling the constructed model to converge more rapidly and potentially achieve superior performance. As a result, the Dice coefficient was increased from 0.60034 to 0.62412.

    The results are as shown in the Table \ref{tab:Model Performance Comparison}.

   \begin{table}[h]
       \centering
       \caption{Model Performance Comparison}
       \label{tab:Model Performance Comparison}
       \begin{tabular}{|c|c|c|}
           \hline
           Model & Train Dice Coefficient & Validation Dice Coefficient \\ \hline
           ResNet18 & 0.59503 & 0.56262 \\ \hline
           ResNet34 & 0.60409 & 0.57165 \\ \hline
           ResNet50 & 0.60034 & 0.59615 \\ \hline
           ResNet101 & 0.57534 & 0.55649 \\ \hline
           ResNet152 & 0.59086 & 0.58404 \\ \hline
           ResNeXt50\_32x4d & 0.60885 & 0.59116 \\ \hline
           ResNeXt101\_32x8d & 0.60349 & 0.58851 \\ \hline
           ResNeSt14d & 0.62412 & 0.62549 \\ \hline
       \end{tabular}
   \end{table}

 \subsection{Experiment 3: Feature Pyramid Network}
 In the subsequent research phase, I successfully implemented a Feature Pyramid Network (FPN) utilizing Shared Memory Parallelism. The foundational work of the FPN architecture, originally proposed by Lin et al\cite{lin2017feature}., consistently outperformed prior models across various performance assessments, including surpassing the unpretrained ResNet50 UNet baseline. FPN's key advantage lies in its capacity to capture features across different scales and integrate them, resulting in the creation of rich multi-scale feature representations. This attribute proves especially advantageous in our application, notably for segmenting aircraft contrails, where the objects of interest may exhibit varying scales within the same image.

 FPN diverges significantly from traditional Convolutional Neural Networks by intentionally generating a multi-scale feature pyramid and conducting final task processing at each pyramid level. This approach integrates top-down pathways and lateral connections into the conventional CNN architecture, departing from the conventional bottom-up feedforward structure found in traditional CNN models. FPN's hierarchical and multi-scale nature positions it as a strong performer in segmentation tasks. These networks are explicitly designed to handle variations in scale and resolution, which is crucial for accurately segmenting intricate and irregular aircraft contrails in high-resolution satellite images.

 The FPN architecture empowers the model to harness semantic features at multiple scales, resulting in enhanced performance in tasks such as object detection. In these tasks, objects may manifest at different scales within the image, and accurate delineation of object boundaries often necessitates a holistic contextual understanding alongside fine-grained details. The FPN model consistently achieved the highest Dice coefficient in all evaluations, reaching an impressive 0.625, even when faced with the challenge of predicting highly irregular objects.

     \begin{figure}[ht]
     \centering
     \includegraphics[width=1\textwidth]{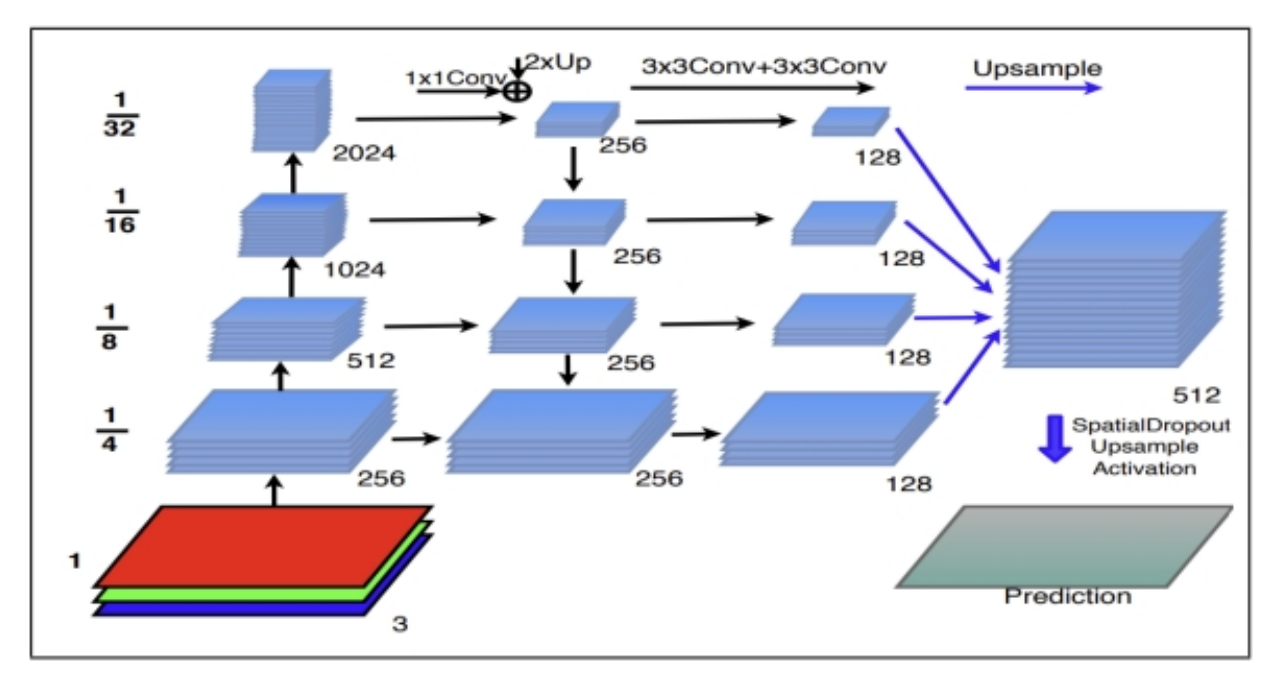}
     \caption{Feature Pyramid Architecture from paper\cite{lin2017feature}.}
     \label{fig:10}
 \end{figure}

 These research findings underscore the immense potential of FPN in image segmentation tasks characterized by multi-scale and resolution variations, providing an effective solution for object detection and segmentation in complex scenes. This study offers valuable insights and serves as a reference point for the future of the image segmentation field.

 \subsection{Experiment 4: Unified Perceptual Parsing Network}
    In the realm of visual perception, humans possess a remarkable multi-level cognitive ability. We can effortlessly classify scenes, detect objects within them, and even recognize intricate details such as textures, surfaces, and the various components of those objects.

    In a groundbreaking research paper, scientists introduced a novel task known as "Unified Perceptual Parsing." This task challenges machine vision systems to comprehensively identify visual concepts from a given image. To tackle this challenge, they introduced a multi-task framework and training strategy called UPerNet\cite{xiao2018unified}, specifically designed to acquire knowledge from diverse image annotations.

    In my own research, I conducted benchmark testing on the Unified Perceptual Parsing task and demonstrated the effectiveness of my framework in segmenting a wide range of visual concepts within images. The trained network exhibited the ability to uncover visual knowledge within natural scenes, allowing it to proficiently segment aircraft contrails and other related visual concepts within images. This capability significantly enhances the model's understanding of complex scenes and its potential applications.

    \subsection{Experiment 5: ConvNeXt Network}
\begin{figure}[ht]
        \centering
        \includegraphics[width=1\textwidth]{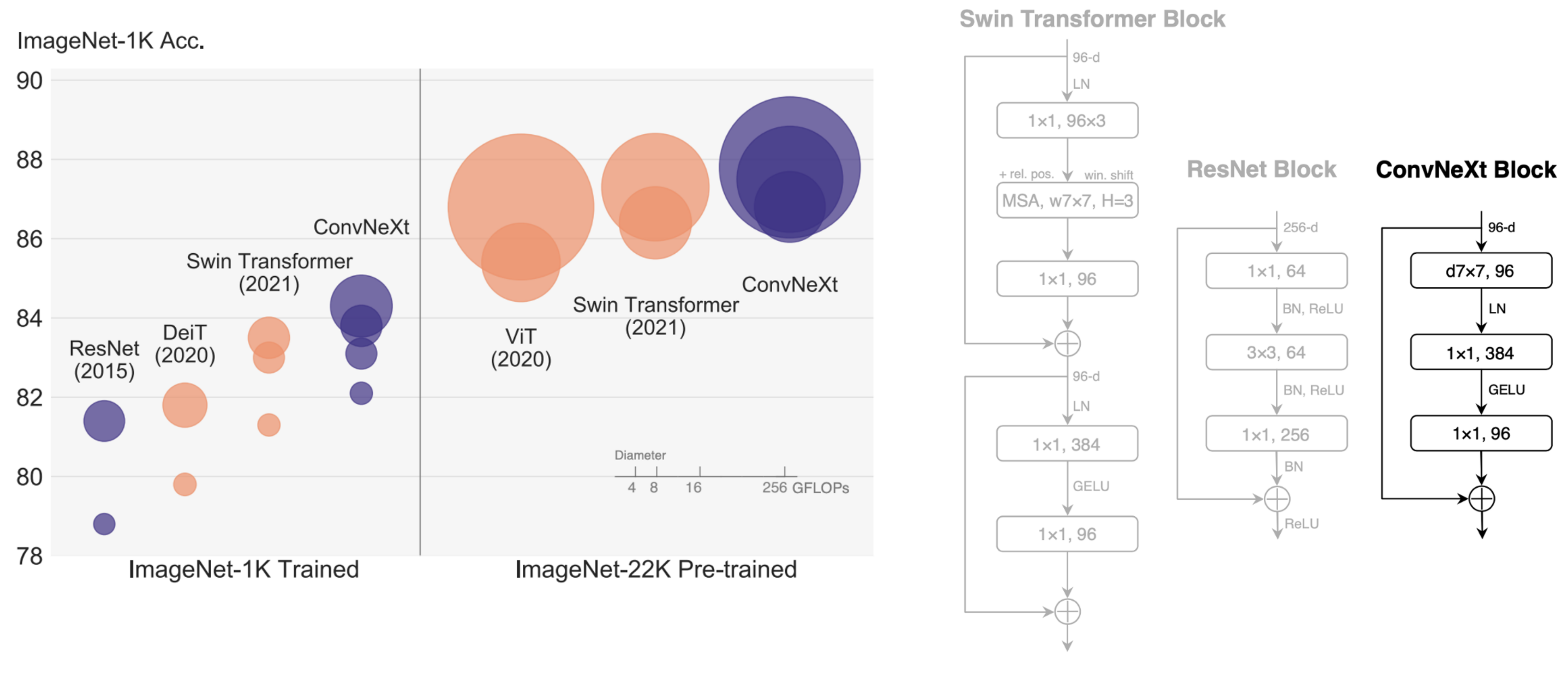}
        \caption{ A pure ConvNet model constructed entirely from standard ConvNet modules. ConvNeXt is accurate, efficient, scalable and very simple in design. \cite{liu2022convnet}}
        \label{fig:11}
    \end{figure}

    The remarkable performance of Transformer and its derivative models in the realms of image classification and recognition has led to a growing trend among researchers. Many are shifting their focus away from traditional CNN-based deep learning models and are increasingly exploring the potential of Transformer networks. Notably, recent collaborative research between Facebook AI Research and UC Berkeley has introduced a groundbreaking convolutional neural network model for the 2020s, known as ConvNeXt\cite{liu2022convnet}. This network has demonstrated exceptional promise, even outperforming the Swin-T model\cite{liu2021swin} in multiple classification and recognition tasks.Some details are as shown in the Figure \ref{fig:11}.

    What distinguishes the ConvNeXt network is that it doesn't introduce significant changes to the overall network architecture. Instead, it ingeniously incorporates advanced concepts and techniques from Transformer networks into the well-established ResNet50/200 networks, thereby enhancing performance. This innovative approach maximizes the strengths of Transformer networks by integrating them with traditional CNN network modules, resulting in a substantial performance enhancement.

    In the contemporary landscape of deep learning, the emergence of the ConvNeXt network represents a forward-thinking approach that harmonizes the advantages of diverse network structures. This synergy has led to notable improvements in image classification and recognition tasks. The findings from this research provide valuable insights for future studies, particularly in tackling complex image tasks and large-scale datasets, where the ConvNeXt network has the potential to be a formidable asset.

    \section{Model}
    \subsection{Train}
    During the model training phase, I leveraged the MMsegmentation library\url{https://github.com/open-mmlab/mmsegmentation}, an open-source semantic segmentation toolkit renowned for its rich model library and versatile training configurations. This toolkit encompasses an extensive model library, encompassing advanced neural network architectures like UNet, DeepLab, and FCN, and offering multiple training configuration options to cater to diverse tasks and datasets. MMsegmentation empowers users to effortlessly construct, train, and fine-tune semantic segmentation models, yielding precise and accurate image segmentation results. Its flexibility and open-source nature have made it a preferred choice across both research and application domains, contributing significantly to the advancements in computer vision.

    For this research, I selected UPerNet as the semantic segmentation model, a fully convolutional network renowned for its end-to-end training and inference capabilities, delivering exceptional performance in research tasks. To enhance the model's capabilities further, I opted for ConvNeXt as the backbone network for UPerNet. ConvNeXt is an innovative convolutional network structure acclaimed for its superior performance. This choice not only bestowed my model with enhanced expressive power but also improved its feature extraction capabilities, a crucial factor in addressing the research task.

    The synergy of UPerNet and ConvNeXt provided a formidable tool for achieving outstanding results in the aircraft contrail semantic segmentation task. This unique model structure, coupled with end-to-end training and inference capabilities, offered a robust solution for the research project, holding significant promise in the field of computer vision.

    When selecting the backbone network, I considered two ConvNeXt variants: ConvNeXt-base and ConvNeXt-large. ConvNeXt-base, a smaller model, is suitable for resource-constrained environments or lightweight tasks while retaining exceptional performance with fewer parameters and computational complexity. In contrast, ConvNeXt-large is a more potent model designed for precision-critical tasks with complex feature extraction requirements. It boasts more network layers and parameters, excelling in large-scale, high-complexity semantic segmentation tasks.

    Based on the specific research requirements, I flexibly chose either ConvNeXt-base or ConvNeXt-large as the model's backbone network, tailoring the selection to the task's needs and optimizing performance within resource constraints. This adaptability allowed for tailored solutions, ensuring top-notch semantic segmentation results across various application scenarios.

    Furthermore, I meticulously defined a comprehensive set of training configurations for the model, encompassing elements like learning rate strategies, optimizer selection, and loss function settings. The optimal learning rate was set at 2.5e-4. For the loss function, I employed the common cross-entropy loss function, suited for image segmentation tasks. To effectively address class imbalance concerns, I thoughtfully assigned class weights, emphasizing rare classes by setting the weight before the positive class to 10.

    The loss function can be expressed as follows, given $N$ samples, $C$ is the number of classes, $p_i$ represents the model's predicted probability distribution, $y_i$ is the true label (0 or 1), and $w$ represents the weight of the positive class (which is 10 times in this case):

    $$
    Loss = -\frac{1}{N}\sum_{i=1}^N\sum_{j=1}^C w_j y_(ij)\cdot \log(p_(ij))
    $$
    Here, $y_{ij}$ is a binary variable that equals 1 if sample i belongs to class j, and 0 otherwise.

This loss function primarily consists of cross-entropy loss, but each sample's contribution is weighted by the corresponding class weight $w_j$, amplifying the contribution of the positive class.

    To further fine-tune model parameters, I selected the AdamW optimizer, a widely-used gradient-based optimization algorithm in deep learning. Additionally, the model incorporated a learning rate decay strategy to ensure training stability while facilitating gradual convergence. This strategy enhanced the model's adaptability to the data and improved its generalization performance.

    Subsequently, I utilized the training functions provided by MMsegmentation to initiate the model training process. During this phase, I adopted a cautious approach, regularly saving model parameters at the end of each epoch to allow for resuming training or performing inference as needed.

    Throughout training, my focus extended beyond achieving model convergence; I conducted extensive evaluations of its performance on the validation dataset. This process served as a vital gauge for monitoring the model's generalization capabilities, ensuring its proficiency on unseen data. Through meticulous hyperparameter tuning and training strategy adjustments, I strived to guarantee the model's exceptional performance on the validation set. This ongoing monitoring and evaluation framework provided timely feedback on training progress, facilitating continuous refinement for optimal semantic segmentation outcomes.

    \subsection{Inference}
    During the model inference phase, I adhered to a meticulously structured series of steps designed to yield high-quality prediction results. To commence, I loaded the meticulously trained model parameters, which were fine-tuned to capture crucial data features. Subsequently, I methodically processed each record in the test set, harnessing the model's predictive capabilities. To bolster the stability and resilience of predictions, I conducted parallel predictions employing multiple models and subsequently averaged the predictions from each model. This approach served to mitigate model variance and elevate overall prediction performance.

    In the course of the prediction procedure, I introduced a threshold, thoughtfully set at 0.75. This threshold played a pivotal role in labeling pixels with probabilities surpassing the 0.75 mark as contrails, thereby generating a binary prediction mask.

    In the final phase, I harnessed the Run-Length Encoding (RLE) encoding technique to compactly encode contrail pixels, ensuring an efficient and compact representation of the segmentation results. These encoded results were meticulously preserved in a submission file, aligning the final model output for effortless participation in competitions or seamless integration into subsequent analyses. This rigorous and systematic inference process was instrumental in guaranteeing the usability and reliability of my semantic segmentation results, making them suitable for practical applications.

    The culmination of these efforts yielded an impressive result of 0.70165, a testament to the effectiveness of the approach and the model's performance.

    Our predicted segmentation results are shown in the following figure \ref{fig:12}.

   \begin{figure}[ht]
    \centering
    \begin{minipage}{1\textwidth}
        \centering
        \includegraphics[width=1\textwidth]{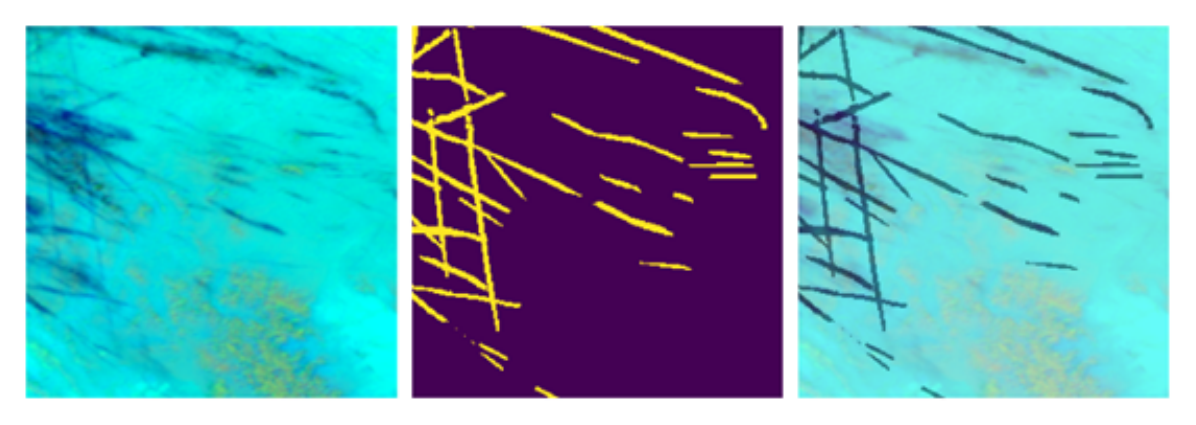}
    \end{minipage}\hfill
    \begin{minipage}{1\textwidth}
        \centering
        \includegraphics[width=1\textwidth]{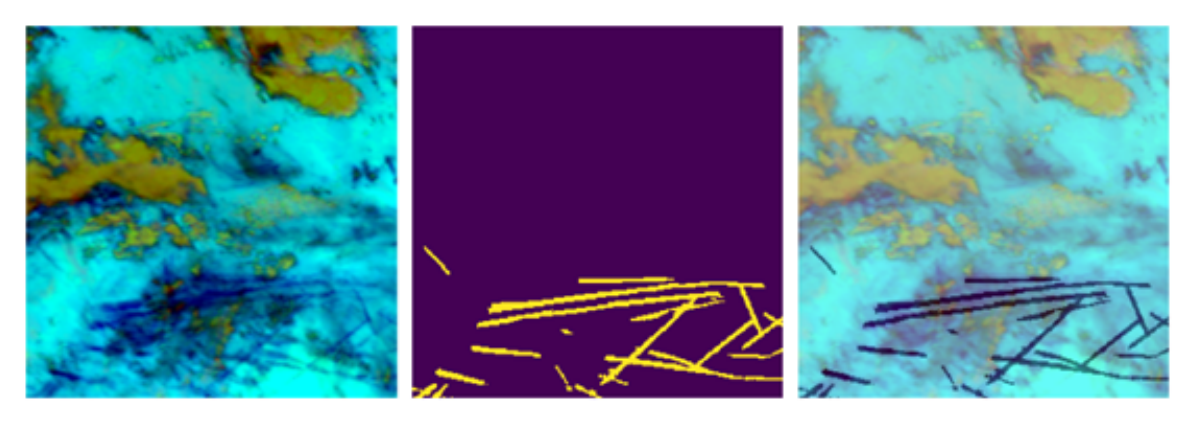}
    \end{minipage}\hfill
    \begin{minipage}{1\textwidth}
        \centering
        \includegraphics[width=1\textwidth]{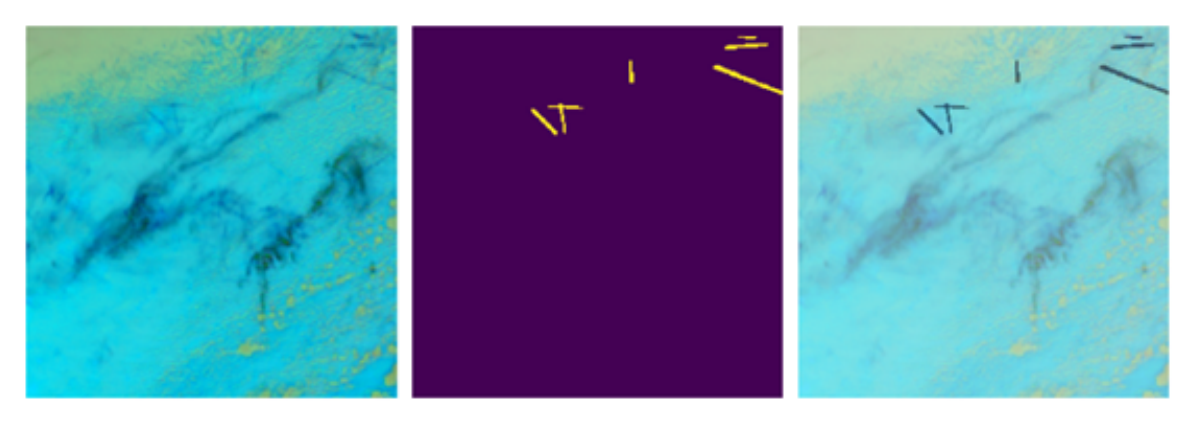}
    \end{minipage}
    \caption{False color image/Ground truth contrail mask/Contrail mask predictions on false color image.}
    \label{fig:12}
\end{figure}

    \section{Conclusion}
    In the aircraft contrails recognition competition hosted by Google Research on global satellite imagery, I meticulously designed and implemented a comprehensive solution that yielded satisfying results.

    Firstly, in the data preprocessing phase, I employed custom functions to extract brightness temperature information from the infrared channel and generated false-color images to enhance the model's perceptual capabilities. This innovative approach enabled the model to better understand the information within the images, resulting in improved recognition performance.

    Given the class imbalance issue, I chose to train the model only on images labeled with positive class instances and fine-tuned the dataset split through 5-fold cross-validation to ensure balanced training data.

    Regarding model selection, I utilized the MMsegmentation library, selecting UPerNet as the foundational framework and incorporating two configurations of ConvNeXt to enhance model performance and robustness. To address the class imbalance effectively, I employed the cross-entropy loss function and introduced weights for the positive class to prioritize accurate recognition of contrails. Additionally, I chose the AdamW optimizer and set the learning rate to 2.5e-4 to optimize model parameters.

    During the model inference phase, I employed a fusion strategy with multiple model predictions and set a threshold of 0.75 for contrail determination to obtain a binary prediction mask. Finally, to organize the prediction results into a submission-ready format, I used the Run-Length Encoding (RLE) encoding method, providing an efficient and reliable output solution for the project.

    Ultimately, the solution achieved excellent Dice coefficient scores, reaching 0.70165 and 0.69445 on two separate datasets, placing me among the top 5\% of outstanding solutions.

\bibliographystyle{plain}
\bibliography{kuan} 

\end{document}